\def\year{2022}\relax
\documentclass[letterpaper]{article} 
\usepackage{aaai22}  
\usepackage{times}  
\usepackage{helvet}  
\usepackage{courier}  
\usepackage[hyphens]{url}  
\usepackage{graphicx} 
\usepackage{natbib}  
\usepackage{caption} 
\DeclareCaptionStyle{ruled}{labelfont=normalfont,labelsep=colon,strut=off} 
\usepackage{algorithm}
\usepackage{algorithmic}
\usepackage{tcolorbox}
\usepackage{enumitem}
\usepackage{multirow}
\usepackage{amsmath,amssymb,amsfonts}
\usepackage{bbold}
\usepackage[switch]{lineno}

\newcount\DraftStatus  
\usepackage{color}
\definecolor{darkgreen}{rgb}{0,0.5,0}
\definecolor{purple}{rgb}{1,0,1}
\definecolor{todocolor}{rgb}{0.9,0.1,0.1}
\definecolor{hycolor}{rgb}{0.7,0.7,0.3}
\definecolor{fixcolor}{rgb}{0.1,0.7,0.3}

\newcommand{\draftnote}[2]{\ifnum\DraftStatus=1
	\marginpar{
		\tiny\raggedright
		\hbadness=10000
		\def\baselinestretch{0.8}
		\textcolor{#1}{\textsf{\hspace{0pt}#2}}}
	\fi}

\usepackage{xspace}
\newcommand{\ourapproach}{\textsc{SynCoBERT}\xspace}

\usepackage{newfloat}
\usepackage{listings}
\floatstyle{ruled}
\newfloat{listing}{tb}{lst}{}
\floatname{listing}{Listing}
\title{\ourapproach: Syntax-Enhanced Contrastive Pre-Training for Code Representation}
\title{\ourapproach: Syntax-Guided Multi-Modal Contrastive Pre-Training for Programming Language Understanding}
\title{\ourapproach: Multi-Modal Contrastive Pre-Training with Syntactical Guidance for Programming Language Understanding}
\title{\ourapproach: Syntax-Guided Multi-Modal Contrastive Pre-Training for Programming Languages}
\title{\ourapproach: Syntax-Guided Multi-Modal Contrastive Pre-Training for\\Code Representation}

\author{Xin Wang\textsuperscript{\rm 1}, Yasheng Wang\textsuperscript{\rm 2}, Fei Mi\textsuperscript{\rm 2}, Pingyi Zhou\textsuperscript{\rm 2}\\ Yao Wan\textsuperscript{\rm 3}, 
	Xiao Liu\textsuperscript{\rm 4}, Li Li\textsuperscript{\rm 5},  Hao Wu\textsuperscript{\rm 6}, Jin Liu\textsuperscript{\rm 1}, Xin Jiang\textsuperscript{\rm 2}}
\affiliations{\textsuperscript{\rm 1}Wuhan University,\ \ 
	\textsuperscript{\rm 2}Noah’s Ark Lab, Huawei\\
	\textsuperscript{\rm 3}Huazhong University of Science and Technology\\
	\textsuperscript{\rm 4}Deakin University,\ \ 
	\textsuperscript{\rm 5}Monash University,\ \ 
	\textsuperscript{\rm 6}Yunnan University\\
	
	\{xinwang0920, jinliu\}@whu.edu.cn,\ \ \{wangyasheng, mifei2, zhoupingyi, Jiang.Xin\}@huawei.com,\\
	wanyao@hust.edu.cn,\ \
	xiao.liu@deakin.edu.au,\ \ 
	Li.Li@monash.edu,\ \ 
	haowu@ynu.edu.cn
}

\usepackage{bibentry}

\begin{document}
	\maketitle
	
	\begin{abstract}
		Code representation learning, which aims to encode the semantics of source code into distributed vectors, 
		plays an important role in recent deep-learning-based models for code intelligence.
		Recently, many pre-trained language models for source code (e.g., CuBERT and CodeBERT) have been proposed to model the context of code and serve as a basis for downstream code intelligence tasks such as code search, code clone detection, and program translation.
		Current approaches typically consider the source code as a plain sequence of tokens, or inject the structure information (e.g., AST and data-flow) into the sequential model pre-training.
		To further explore the properties of programming languages, this paper proposes \ourapproach, a \underline{Syn}tax-guided multi-modal contrastive pre-training approach for better \underline{Co}de representations.
		Specially, we design two novel pre-training objectives originating from the symbolic and syntactic properties of source code, i.e., \textit{Identifier Prediction (IP)} and \textit{AST Edge Prediction (TEP)}, which are designed to predict identifiers, and edges between two nodes of AST, respectively.
		Meanwhile, to exploit the complementary information in semantically equivalent modalities (i.e., code, comment, AST) of the code, we propose a multi-modal contrastive learning strategy to maximize the mutual information among different modalities.
		Extensive experiments on four downstream tasks related to code intelligence show that \ourapproach advances the state-of-the-art with the same pre-training corpus and model size.
	\end{abstract}
	
	\section{Introduction}
	Code intelligence that utilizes machine learning techniques to promote the productivity of software developers, has attracted increasing interest in both communities of software engineering and artificial intelligence~\cite{lu2021codexglue}.
	To achieve code intelligence, one fundamental task is code representation learning (also known as code embedding), which can support many downstream tasks, including code search~\cite{gu2018deep},
	code clone detection~\cite{white2016deep}, code defect detection~\cite{li2017software}, and program translation~\cite{chen2018tree}.
	
	\begin{figure}
		\centering
		\includegraphics[width=0.42\textwidth]{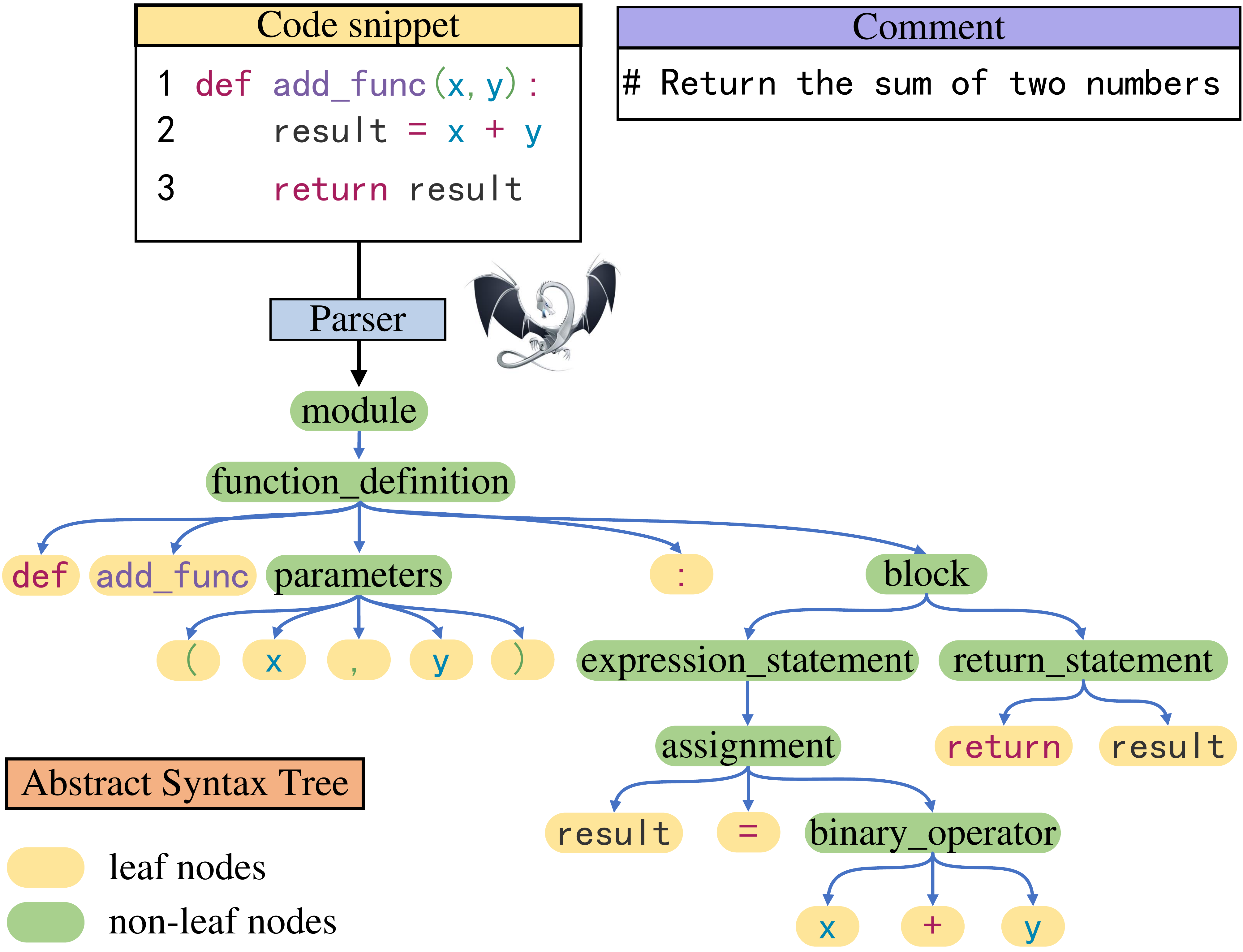}
		\caption{\label{fig:ast} 
			A Python code snippet with its AST.
		}
		\vspace{-1em}
	\end{figure}
	
	Benefiting from the strong power of pre-training techniques in natural language processing (NLP), such as BERT \cite{devlin2018bert}, RoBERTa~\cite{liu2019roberta}, ALBERT~\cite{lan2020albert} and GPT~\cite{radford2018improving}, there are several attempts to pre-train a language model on the corpus of source code for code understanding and generation. 
	CodeBERT~\cite{feng2020codebert} is a bimodal pre-trained model on the combination of source code and natural-language descriptions.
	PLBART~\cite{ahmad2021unified} is a unified pre-training framework for code understanding and generation.
	To incorporate the syntax structure of code, \citet{guo2021graphcodebert} further propose GraphCodeBERT to preserve the syntax structure of source code by introducing an edge masking technique over data-flow graphs.
	\citet{jiang2021treebert} propose TreeBERT, 
	which incorporates the abstract syntax tree (AST) of the code into model pre-training by designing two objectives of tree-based masked language modeling and node order prediction.
	
	Despite much progress having been made towards pre-training language models for source code, several characteristics of programming languages are still not sufficiently explored.
	To better represent the syntax structure of code, we consider two crucial but overlooked characteristics of source code.
	\textbf{(1)}. \textit{Code identifier contains symbolic and syntactic information.} \textit{Identifier} or variable is a basic component for programming language. It should not be simply considered as regular textual code tokens. For example, given an expression \texttt{x = len("x")}, the previous \texttt{x} is a \textit{identifier}, which differentiates the later string \texttt{x}.
	\textit{identifiers} play an important role in understanding the logic of code, because they contain crucial symbolic and syntactic information.
	\textbf{(2)}. \textit{The syntax information along AST edges is ignored.} 
	To obtain the syntax structure of code, AST has been widely adopted~\cite{zugner2021language,jiang2021treebert}. Figure~\ref{fig:ast} shows a Python code snippet with its AST. 
	In this AST, a binary operator statement \texttt{x + y} can be represented by a \textit{non-leaf} node \texttt{binary\_operator} points to three \textit{leaf} nodes (\texttt{x}, \texttt{y}, and an operational character \texttt{+}). We argue that the edges connecting \textit{non-leaf} nodes and \textit{leaf} nodes contain rich syntactic structure information that should be considered.
	
	In addition, a program is often composed of both code snippets and corresponding comments. 
	Furthermore, a code snippet can be parsed into one or more syntactic structures (e.g., AST or control-/data- flow graph). 
	In this paper, we call these code features from different perspectives as multiple modalities of code.
	We argue that these semantically equivalent modalities provide complementary information to learn more comprehensive code representations. 
	However, previous works do not further explore the potential mutual information among different modalities of the code.

	Motivated by the aforementioned limitations, this paper proposes \ourapproach, a \underline{Syn}tax-guided multi-modal contrastive pre-training approach for better \underline{Co}de representations.
	We design two novel pre-training objectives originating from the symbolic and syntactic properties of source code, i.e., \textit{Identifier Prediction (IP)} and \textit{AST Edge Prediction (TEP)}, which are designed to predict identifiers, and edges between two nodes of AST, respectively.
	We propose a \textit{multi-modal contrastive learning (MCL)} objective to obtain more comprehensive representations by learning from three modalities (code, comment, and AST) through contrastive learning.
	Overall, the key contributions of this paper are as follows:
	\begin{itemize}[itemsep=-1pt,topsep=0pt,leftmargin=*]
		\item We propose \ourapproach, a syntax-guided multi-modal contrastive pre-training framework for code representations. We design two new pre-training objectives to encode the symbolic and syntactic information of programming languages. The first IP objective predicts whether the code token is an identifier or not. The second TEP objective predicts the edges between two nodes of AST.
		
		\item We propose a multi-modal contrastive pre-training strategy that maximizes mutual information among different modalities (code, comment, and AST) through contrastive learning to learn more comprehensive representations.
		
		\item Comprehensive experiments conducted on four code intelligence tasks (code search, code clone detection, code defect detection, program translation) demonstrate that \ourapproach\ advances the state-of-the-art with the same pre-training corpus and model size.
	\end{itemize}
	
	\section{\ourapproach}
	This section first gives preliminaries about input representations and model architecture of \ourapproach. Then, we introduce our novel pre-training framework covering four objectives, including Multi-Modal Masked Language Modeling (MMLM), Identifier Prediction (IP), AST Edge Prediction (TEP), and Multi-Modal Contrastive Learning (MCL).
	
	\begin{figure}
		\centering
		\includegraphics[width=0.48\textwidth]{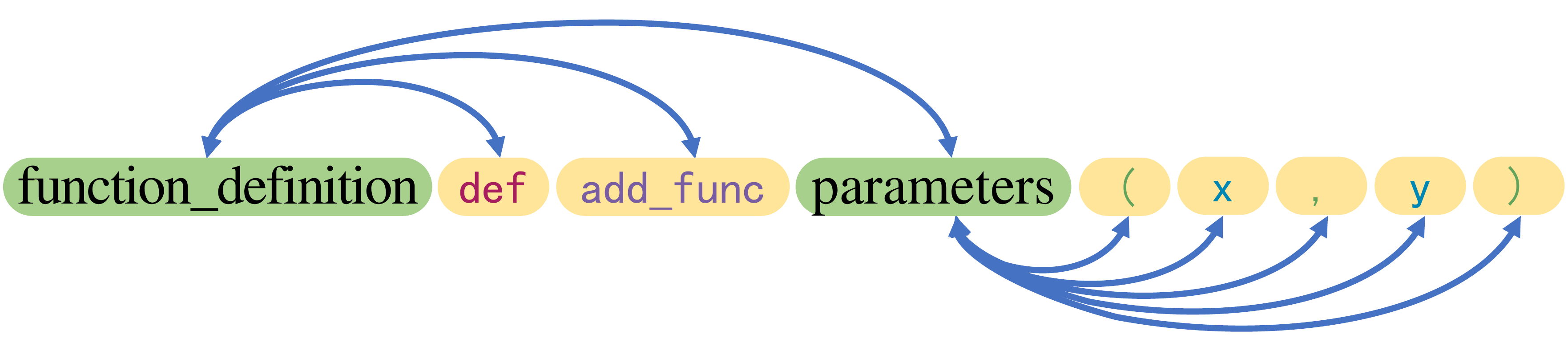}
		\caption{\label{fig:parse} A part of the AST sequence obtained from the AST in Figure~\ref{fig:ast}, blue arrows denote edges between nodes.} 
		\vspace{-1em}
	\end{figure}
	
	\begin{figure*}
		\centering
		\includegraphics[width=0.96\textwidth]{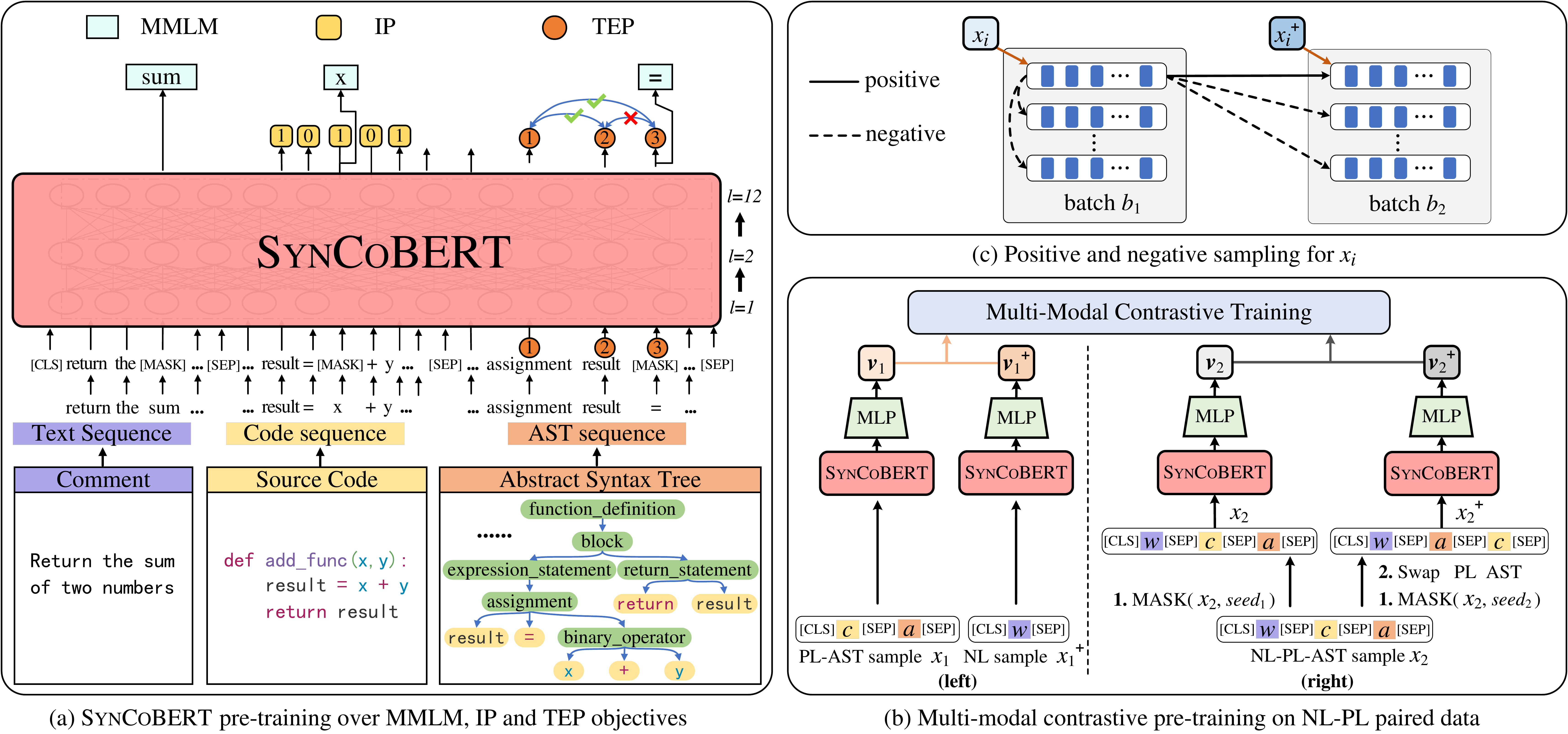}
		\caption{\label{fig:model}  Different scenes of \ourapproach pre-training. (a) \ourapproach takes source code paired with comment and the corresponding AST as the input, and is pre-trained with MMLM, IP, TEP objectives. (b) 
			Positive sampling for NL-PL paired data, (left) NL \textit{vs} PL-AST, (right) NL-PL-AST \textit{vs} NL-AST-PL. (c) An illustration about positive and negative pairs, including \textit{in-batch} and \textit{cross-batch} negative sampling.
		}
		\vspace{-1em}
	\end{figure*}
	
	\subsection{Preliminary}
	\paragraph{Input Representations}
	We take the AST of the code as part of the input to the model, which provides an AST token sequence with a depth-first traversal. We adopt tree-sitter\footnote{https://github.com/tree-sitter/tree-sitter} to convert the code into an AST. Figure \ref{fig:parse} shows an example of partial AST sequence obtained from the AST in Figure \ref{fig:ast}, and the blue arrows denote edges between nodes.
	Given a natural language comment $w=\{w_1, w_2, \ldots,  w_{|w|} \}$, the corresponding source code $c = \{c_1,c_2,\ldots,c_{|c|}\}$, and the AST sequence $a =\{a_1,a_2,\ldots,a_{|a|}\}$,
	\ourapproach takes the concatenation of multiple modalities (NL, PL, AST) as input, i.e., 
	\begin{equation}
	x=\{{\texttt{[CLS]}}, w,{\texttt{[SEP]}}, c,{\texttt{[SEP]}}, a, {\texttt{[SEP]}} \},
	\end{equation}
	where $\texttt{[CLS]}$~\cite{devlin2018bert} is a special token for ``classification tasks'', appearing at the beginning of the input sequence, and $\texttt{[SEP]}$ is a special token to split two kinds of sub-sequences. For unimodal code data (without NL comments), the input is $x=\{{\texttt{[CLS]}}, c,{\texttt{[SEP]}}, a, {\texttt{[SEP]}} \}$.
	
	\paragraph{Model Architecture}
	\ourapproach is built on a multi-layer Transformer~\cite{vaswani2017attention} encoder as in BERT, which we will not review in detail. 
	The embedding of each token in $x$ is the sum of corresponding token and position embeddings. \ourapproach adopts 12-layer Transformer with 768 hidden sizes and 12 attention heads to encode the input vectors into contextual representations.

	\subsection{Multi-Modal Masked Language Modeling (MMLM)} Our approach jointly models NL, PL, and AST, providing complementary information contained in multiple modalities. We extend the NLP task know as Masked Language Model (MLM)~\cite{devlin2018bert} to multiple modalities.
	Given a data point of NL-PL-AST triplet $\{{w, c, a}\}$ as input, we randomly select 15\% of tokens from the concatenation of NL, PL, and AST. Following the same settings in~\cite{devlin2018bert}, we replace 80\% of them with \texttt{[MASK]} tokens, 10\% with random tokens, and the remaining 10\% unchanged. 
	The MMLM objective is a cross-entropy loss on predicting the masked tokens as: 
	\begin{equation}
	\small
	\mathcal{L}_{\rm MMLM} = -\sum^V \sum_{i}^{M} y_i^{\rm MMLM} \ {\rm ln} p_i^{\rm MMLM}\,,
	\end{equation}
	where $M = w^m\cup c^m\cup a^m$ is the set of masked tokens for NL ($w^m$), PL ($c^m$) and AST ($a^m$).
	$V$ represents the vocabulary size, $y_i^{\rm MMLM}$ denotes the label of the masked token $i$, and $p_i^{\rm MMLM}$ is the predicted probability of token $i$. 
	
	\subsection{Identifier Prediction (IP)}
	Previous works often overlook the symbolic property of programming languages. 
	As a typical symbol, the \textit{identifier} plays an important role in source codes. It can be replaced by another string without affecting the logic of the source code. 
	As it is prohibitive the exhaustively predict a large number of code token types in source code, we only divide the code token types into \textit{identifier} or \textit{non-identifier}, considering the importance and large proportion of identifiers.
	Different from MMLM (predicting 15\% code tokens), we pose the identifier prediction objective over all code tokens.
	For each token in the source code, a label 1 is applied if it is an \textit{identifier}, and a label 0 is applied otherwise (c.f. Figure~\ref{fig:model}(a)).
	Therefore, the IP loss function is a binary classification loss defined as:
	\begin{equation}
	\small
	\label{eq:type} 
	\mathcal{L}_{\rm IP}\! =\! -\sum_{i\in c}[y_{i}^{\rm IP} {\rm ln} p_{i}^{\rm IP}\! +\! (1-y_{i}^{\rm IP}){\rm ln}(1-p_{i}^{\rm IP})]\,,
	\end{equation}
	
	\noindent where $p_i^{\rm IP}$ is the predicted identifier probability of the $i$-th code token, and
	$y_i^{\rm IP}$ is the label of the $i$-th code token.
	
	\subsection{AST Edge Prediction (TEP)}
	When converting an AST tree into a sequence, some crucial structural information might get lost. Some existing studies, such as \textit{Tree Transformer}~\cite{wang2019tree} and \textit{Tree LSTM}~\cite{tai2015improved}, put trees into models by introducing additional modules. Inspired by the edge masking technique over data-flow graphs proposed in GraphCodeBERT~\cite{guo2021graphcodebert}, we design an AST edge prediction objective to encode the tree structure information into the model simply and directly without introducing additional modules.
	Taking the token ``\texttt{result}'' as an example in Figure~\ref{fig:model}(a), there is an edge between tokens (``\texttt{assignment}'', ``\texttt{result}''), and there is no edge between (``\texttt{result}'', ``\texttt{=}''). 
	To incorporate such tree structural information, we mask edges in the AST and ask the model to predict these edges. Formally, the loss function of this TEP objective is defined as:
	\begin{equation}
	\small
	\label{eq:bce} 
	\mathcal{L}_{\rm TEP}\!=\!-\!\sum_{(i,j)\in N_a}[y_{(i,j)}^{\rm TEP} {\rm ln} p_{(i,j)}^{\rm TEP} + (1-y_{(i,j)}^{\rm TEP}){\rm ln}(1-p_{(i,j)}^{\rm TEP})]\,,
	\end{equation}
	
	\noindent where $N_a$ represents the set of all AST node pairs. $y_{(i,j)}^{\rm TEP}$ is 1 if there is an edge between the $i$-th and $j$-th nodes, otherwise $y_{(i,j)}^{\rm TEP}=0$. $p_{(i,j)}^{\rm TEP}$ is the probability of whether there is an edge between the $i$-th and $j$-th nodes, which is calculated by dot product using representations of these two nodes. A $\operatorname{sigmoid}$ activation function is utilized to normalize the value of $p_{(i,j)}^{\rm TEP}$ within the range of 0 to 1.
	
	\subsection{Multi-Modal Contrastive Learning (MCL)}
	
	Previous works~\cite{li2020on, reimers2019sentence} have shown that native sentence representations derived from BERT are dominated by high-frequency tokens.
	This \textit{token imbalance} issue is even more serious in codes. Taking the Python language as an example, the ``\texttt{def}'' token appears in almost all functions. 
	Contrastive learning encourages the representation of the original sequence to be closer to the representation of the ``positive'' augmented sequence, while staying away from representations of ``negative'' sequences, making the model learn a more even decision boundary across different data points to reconcile the representation bias caused by token imbalance~\cite{yan2021consert}.
	Several recent works~\cite{jain2021contrastive, bui2021self} attempt to compare similar and dissimilar code snippets. However, they only handle the single modality of code, and ignore the multi-modal characteristic of programming languages. These semantically equivalent modalities can provide complementary information to learn more comprehensive code representations.
	To this end, we propose a Multi-Modal Contrastive Learning (MCL) objective to explore the potential for maximizing mutual information among different modalities, encouraging the model to learn useful connections and semantic equivalences.
	
	To be specific, we train \ourapproach with \textit{paired data} and \textit{unpaired data}. Paired data refers to codes (PL) with paired natural language comments (NL), while unpaired data stands for standalone codes without paired natural language comments. 
	Next, we explain how we construct positive and negative samples for these two cases.
	
	\paragraph{Positive Samples}
	For paired data, we design two simple and effective ways to construct positive samples for MCL:
	\begin{itemize}[leftmargin=*]
		\item \textbf{NL} \textit{vs} \textbf{PL-AST.}  To bridge the gap between a natural language comment with its code snippet, we consider a comment (NL) as a positive sample w.r.t the corresponding code and AST. That it, \textbf{NL \& PL-AST} forms a positive pair, and an illustration example can be found at $x_1$ and $x_1^+$  of Figure~\ref{fig:model}(b)(left). 
		\item \textbf{NL-PL-AST} \textit{vs} \textbf{NL-AST-PL.}  To better learn the semantic equivalence between PL and AST conditioned on the same NL comment, we propose to construct another group of positive samples by reversing the order of PL ($c$) and AST ($a$) in the input triplet $\{w, c, a\}$ to be $\{w, a, c\}$. Before this order swapping operation, the original input triplet is firstly masked by different random seeds. This step aims to increase differences between positive pairs.  A concrete example of these steps is illustrated in Figure~\ref{fig:model}(b)(right). 
	\end{itemize}
	For unpaired data, we construct positive sample pairs considering \textbf{PL-AST} \textit{vs} \textbf{AST-PL.} This scheme works the same as the setting of \textbf{NL-PL-AST} \textit{vs} \textbf{NL-AST-PL} introduced above without considering the NL comment.

	\paragraph{Negative Samples}
	To obtain negative samples for MCL, we adopt \textit{in-batch} and \textit{cross-batch} sampling during training. For a batch of training data $b_1 = [x_1 \dots x_N]$ with size $N$, we can first obtain another positive data batch $b_2 = [x_1^+ \dots x_N^+]$ with size $N$ using schemes described before where $\{x_i, x_i^+\}$ is a positive pair. 
	For $x_i$, the \textit{in-batch} and the \textit{cross-batch} negative samples are $\{x_j\}, \forall {j \neq i}$, 
	In this way, we can obtain a set $\mathbf{X}^-$ of $2N-2$ negative samples for each $x_i$, and an illustrative example is given in Figure~\ref{fig:model}(c).

	\paragraph{Overall MCL Pipeline}  For an input $x_i$ in paired data,  
	the following steps are executed (unpaired data is similar):
	\begin{itemize}[leftmargin=*]
		\item First, we construct a positive sample $x_i^+$ for $x_i$, and the two ways to construct positive samples for paired data introduced before are illustrated in Figure~\ref{fig:model}(b).
		\item Second, we take $x_i$ and $x_i^+$ as inputs of \ourapproach. Then we can obtain the vector representations of them $\boldsymbol h_i = {\ourapproach(x_i)}$ and $\boldsymbol h_i^+ = {\ourapproach(x_i^+)}$.
		\item Finally, we adopt a two-layer MLP 
		$f(\cdot)$ that maps representations to the space $\boldsymbol v_i = f(\boldsymbol h_i), \boldsymbol v_i^+ = f(\boldsymbol h_i^+)$ where contrastive loss is applied. Through the nonlinear transformation, more information can be maintained in $\boldsymbol h$~\cite{chen2020a}.
	\end{itemize}
	
	For an input $x_i$ with representation $\boldsymbol v_i$, it has one positive sample $x_i^+$ with representation $\boldsymbol v_i^+$, and it also has a set of negative samples $\mathbf{X}^-$ with size $2N-2$. We denote the set of representations for samples in $\mathbf{X}^-$ as $\mathbf{V}^- = \{\boldsymbol{v}_1^-, \dots, \boldsymbol{v}_{2N-2}^-\}$. The objective of contrastive learning is to maximize the representation similarity between positive samples, while minimizing the representation similarity between negative samples.
	Therefore, we define the loss function for a positive pair ($x_i,x_i^+$) as:
	\begin{equation}
	\small
	\label{eq:cl} 
	l(x_i,x_i^+) = - {\rm ln} \frac{{\rm exp}(\boldsymbol v_i \cdot \boldsymbol{v}_i^+)}
	{
		{\rm exp}(\boldsymbol v_i \cdot \boldsymbol{v}_i^+) + \sum_{k=1}^{2N-2}{\rm exp}(\boldsymbol{v}_i \cdot \boldsymbol{v}_k^-)
	}\,,
	\end{equation}
	where the similarity of a pair of samples is defined by the dot product of their representations as: $\boldsymbol v_i \cdot \boldsymbol v_j$. 
	This loss has been used in previous works \cite{chen2020a, wu2020clear}. 
	We calculate the loss for the same pair twice with order switched, i.e., ($x_i,x_i^+$) changes to ($x_i^+,x_i$) as the dot product with negative samples for $x_i$ and $x_i^+$ are \textit{different}. Finally, the overall MCL loss is defined as follows:
	\begin{equation}
	\small
	\label{eq:batch_cl} 
	\mathcal{L}_{\rm MCL}= \sum^{N}_i \left[ l(x_i,x_i^+) + l(x_i^+,x_i) \right]\,,
	\end{equation}

	
	\subsection{Training Objective}
	The overall loss function in \ourapproach is the integration of the several components we defined before as:
	\begin{equation}
	\small
	\mathcal{L} = \mathcal{L}_{\rm MMLM} + \mathcal{L}_{\rm IP} + \mathcal{L}_{\rm TEP} + \mathcal{L}_{\rm MCL} + \lambda \lVert \Theta\rVert^2\,,
	\end{equation}
	where $\Theta$ contains all trainable parameters of the model. $\lambda$ is the $L_2$ regularization coefficient used to prevent overfitting.
	
	\begin{table*}[htbp]
		\normalsize
		\centering
		\setlength{\tabcolsep}{2.3mm}{
			\caption{Results on the natural language code search task evaluating with MRR, using the AdvTest and CodeSearch datasets. 
			} 
			\label{table:codesearch}
			\begin{tabular}{l|c|c c c c c c c}
				\hline
				\multirow{2}{*}{Model}&AdvTest&\multicolumn{7}{|c}{CodeSearch}\\
				\cline{2-9}
				&Python & Ruby &Javascript&Go &Python &Java &PHP &Average \\
				\hline
				NBow&- &16.2 &15.7 &33.0 &16.1 &17.1 &15.2 &18.9\\
				CNN &- &27.6 &22.4 &68.0 &24.2 &26.3 &26.0 &32.4\\
				BiRNN&-&21.3 &19.3 &68.8 &29.0 &30.4 &33.8 &33.8\\
				Transformer&-&27.5&28.7&72.3 &39.8 &40.4 &42.6 &41.9\\
				\hline
				RoBERTa&18.3&58.7&51.7&85.0&58.7&59.9&56.0&61.7\\
				RoBERTa (code)&-&62.8&56.2&85.9&61.0&62.0&57.9&64.3\\
				CodeBERT&27.2&67.9&62.0&88.2&67.2&67.6&62.8&69.3\\
				GraphCodeBERT&35.2&70.3&64.4&89.7&69.2&69.1&64.9&71.3\\
				\ourapproach&\textbf{38.1}&\textbf{72.2}&\textbf{67.7}&\textbf{91.3}&\textbf{72.4}&\textbf{72.3}&\textbf{67.8}&\textbf{74.0}\\
				\hline
		\end{tabular}}
	\end{table*}
	\section{Experiment Setup}
	
	\subsection{Pre-Training Dataset and Settings}
	
	For fair comparisons, we pre-train \ourapproach on the CodeSearchNet dataset~\cite{husain2019codesearchnet}, which is the same as that used by CodeBERT and GraphCodeBERT. CodeSearchNet dataset contains 2.1M bimodal data points (code functions paired with natural language comments) and 6.4M unimodal codes across six programming languages.
	More details are presented in the supplementary materials.
	
	We train \ourapproach using Transformer with 12 layers, 768 hidden sizes, and 12 attention heads. \ourapproach is trained on 8 NVIDIA Tesla V100 with 32GB memory. The lengths of sequences containing special tokens in NL, PL, and AST are set to 96, 160, and 256, respectively. The batch size is set to 128. The learning rate is set to 1e-4. We use an Adam optimizer to optimize the parameters of the model. We employ a Byte-Pair Encoding (BPE) tokenizer \cite{sennrich2016neural}.
	To accelerate the process of training, we adopt the parameters of CodeBERT to initialize the model same as GraphCodeBERT.
	Finally, the model is trained with 110K steps and costs about 80 hours.
	
	
	
	\subsection{Evaluation Tasks, Datasets, and Metrics}
	Since \ourapproach belongs to the BERT-like code pre-trained models, it is more suitable for programming language understanding (PLU) tasks, so we choose all three PLU tasks in CodeXGLUE~\cite{lu2021codexglue}, including natural language code search, code clone detection, and code defect detection. To reflect the generality of the model, we also select a programming language generation (PLG) task, such as program translation. 
	
	\paragraph{Natural Language Code Search} is to match the most semantically relevant code functions through natural language queries. We use the AdvTest dataset~\cite{lu2021codexglue} to conduct the experiment on Python language. In order to evaluate other programming languages, we also adopt CodeSearch dataset ~\cite{guo2021graphcodebert}, including six programming languages (Ruby, Javascript, Go, Python, Java, PHP). We adopt Mean Reciprocal Rank (MRR) to evaluate the performances of all code search methods. 
	
	\paragraph{Code Clone Detection} is to identify the existence of code clone issues by measuring the similarity between two code snippets. We fine-tune \ourapproach on the BigCloneBench~\cite{svajlenko2014towards} and POJ-104~\cite{mou2016convolutional} datasets. In the BigCloneBench dataset (Java), given two codes, the task is to judge whether they are semantically similar, evaluating by Precision, Recall, and F1-score. In the POJ-104 dataset (C/C++), given a code, the task retrieves 499 codes evaluating by Mean Average Precision (MAP).
	
	\paragraph{Code Defect Detection} is to identify whether it is an insecure code that may attack software systems.
	It can be regarded as a binary classification task.
	We evaluate all models on Defects4J dataset (C language) \cite{zhou2019devign}, using the Accuracy score.
	
	\paragraph{Program Translation} is to translate the code of one programming language into another.
	We adopt CodeTrans dataset~\cite{chen2018tree}, which contains the mutual translation of C\# and Java. All methods are evaluated by BLEU-4, Exact Match, and CodeBLEU~\cite{lu2021codexglue}.
	
	All datasets except CodeSearch are provided in CodeXGLUE~\cite{lu2021codexglue}, and the default training/validation/testing splits are used. All evaluation task datasets and fine-tuning details are presented in the supplementary materials.
	
	\subsection{Baseline Methods}
	We compare \ourapproach with various state-of-the-art methods in two categories. In the first category, models are directly trained on the evaluation task from scratch. In the second category, models are pre-trained on unlabeled corpus first and then fine-tuned on the evaluation task.
	
	\subsubsection{Training from Scratch}
	\begin{itemize}[itemsep=-1pt,topsep=0pt,leftmargin=*]
		\item \textbf{NBow} is short for bag-of-words, which ignores the word order, grammar, syntax, and other elements of the sequence. It selects candidates based on the number of shared works for natural language code search tasks.
		\item \textbf{Naive copy, PBSMT} are used in program translation tasks. \textbf{Naive copy} means copying the source code as the translation result. \textbf{PBSMT}~\cite{koehn2003statistical} is a statistical machine translation method based on phrases. 
		\item \textbf{TextCNN}~\cite{kim2014convolutional} is a CNN-based model to capture the features of NL or PL at the word level.
		\item \textbf{BiLSTM}~\cite{cho2014on} is a Seq2Seq model based on bidirectional LSTM with an attention mechanism~\cite{luong2015effective}.
		\item \textbf{Transformer}~\cite{vaswani2017attention} is the base architecture of \ourapproach and other pre-trained models. We use the same number of layers and hidden size as pre-trained models.
	\end{itemize}
	\subsubsection{Pre-Trained Models}
	\begin{itemize}[itemsep=-1pt,topsep=0pt,leftmargin=*]
		\item \textbf{RoBERTa}~\cite{liu2019roberta} is pre-trained on natural languages.
		\item \textbf{RoBERTa(code)} is a varient of RoBERTa, and is pre-trained on source code from CodeSearchNet corpus.
		\item \textbf{code2vec}~\cite{alon2019code2vec} uses a soft-attention mechanism on AST paths of the code, and aggregate all of their vector representations into a single vector. \citet{coimbra2021on} pre-trained the code2vec on open-source C language corpus for code defect detection task.
		\item \textbf{CodeBERT}~\cite{feng2020codebert} is pre-trained on PL-NL pairs with MLM and RTD pre-training objectives.
		\item \textbf{GraphCodeBERT}~\cite{guo2021graphcodebert} is pre-trained on the basis of CodeBERT, integrating data flow of codes.
	\end{itemize}
	
	\section{Results and Analysis}
	\subsection{Natural Language Code Search}
	Table~\ref{table:codesearch} summarizes the results of natural language code search on different datasets. 
	The baseline results on the AdvTest dataset are reported in~\cite{lu2021codexglue}. We leverage the checkpoint of GraphCodeBERT to obtain the result on the AdvTest dataset.
	Baseline results on the CodeSearch dataset are all reported in~\cite{guo2021graphcodebert}. 
	
	We can observe that \ourapproach outperforms \textit{all} methods on two datasets. On average, it outperforms CodeBERT by 10.9 points on the AdvTest dataset and 4.7 points on the CodeSearch dataset.
	Compared to GraphCodeBERT, it scores 2.9 points higher on the AdvTest dataset, and 2.7 points averagely higher on the CodeSearch dataset. \ourapproach is pre-trained on the same pre-training corpus as CodeBERT and GraphCodeBERT. The significant performance improvement indicates that \ourapproach can learn better code representations. We credit this improvement to the introduction of multi-modal contrastive learning. Moreover, compared to the Transformer baseline, \ourapproach achieves a 76.6\% improvement on the CodeSearch dataset, and we credit this improvement to the pre-training step.
	\begin{table}[t!]
		\normalsize
		\centering
		\setlength{\tabcolsep}{1mm}{
			\caption{Results on the clone detection task, using the BigCloneBench (Java) and POJ-104 (C/C++) datasets.} 
			\label{table:clone_detection}
			\begin{tabular}{l|c c c|c}
				\hline
				\multirow{2}{*}{Model} &\multicolumn{3}{|c|}{BigCloneBench} &POJ-104\\
				\cline{2-5}
				&Precision&Recall&F1-score&MAP@R\\
				\hline
				RoBERTa&93.5&96.5&94.9&76.67\\
				CodeBERT&96.0&96.9&96.5&82.67\\
				GraphCodeBERT&\textbf{97.3}&96.8&97.1&85.16\\
				\ourapproach&97.1&\textbf{97.7}&\textbf{97.4}&\textbf{88.24}\\
				\hline
		\end{tabular}}
	\end{table}
	
	\subsection{Code Clone Detection}
	Table~\ref{table:clone_detection} shows experimental results on the code clone detection task. The baseline results are reported in~\cite{lu2021codexglue} and~\cite{guo2021graphcodebert}. 
	We can observe that \ourapproach achieves the best overall performance on the two datasets.
	On the BigCloneBench dataset, \ourapproach outperforms all other methods w.r.t. Recall and F1-score although it is slightly worse than GraphCodeBERT w.r.t. Precision. 
	On the POJ-104 dataset, \ourapproach consistently outperforms all methods. Compared to CodeBERT and GraphCodeBERT, it achieves 5.57 and 3.08 points higher respectively. As \ourapproach is not pre-trained on C/C++ languages, the superior performance on the POJ-104 dataset indicates that \ourapproach learns better generic program semantics. These results indicate that \ourapproach indeed learns better code representations for code clone detection.
	
	\subsection{Code Defect Detection}
	The experimental results on code defect detection task are shown in Table~\ref{table:defect_detection}.
	Several representative baseline results are reported in~\cite{lu2021codexglue}. We can observe that \ourapproach outperforms all representative models. Specifically, it outperforms CodeBERT and GraphCodeBERT by 2.42 and 1.29, respectively. The performance improvement indicates that \ourapproach learns more comprehensive program semantics for defect detection. It is also noteworthy that \ourapproach is not pre-trained on C language, while outperforms the code2vec (pre-trained on C language). This result indicates that \ourapproach can learn general syntax information, benefiting other programming languages.
	\begin{table}[t!]
		\normalsize
		\centering
		\caption{Results on the code defect detection task, using the Defects4J dataset (C language).} 
		\label{table:defect_detection}
		\begin{tabular}{l|c}
			\hline
			Models & Accuracy \\
			\hline
			BiLSTM&59.37\\
			TextCNN & 60.69\\
			RoBERTa & 61.05\\
			CodeBERT & 62.08\\
			code2vec & 62.48\\
			GraphCodeBERT& 63.21\\
			\ourapproach& \textbf{64.50}\\
			\hline
		\end{tabular}
	\end{table}

	\begin{table*}[htbp]
		\normalsize
		\centering
		\setlength{\tabcolsep}{2.5mm}{
			\caption{Results on the code translation task with BLEU, Accuracy and CodeBLEU score, using the CodeTrans dataset.}
			\label{table:codetrans}
			\begin{tabular}{l|c c c|c c c}
				\hline
				\multirow{2}{*}{Methods}&\multicolumn{3}{|c}{C\#$\rightarrow$Java}&\multicolumn{3}{|c}{Java$\rightarrow$C\#}\\
				\cline{2-7}
				&BLEU&Exact Match&CodeBLEU&BLEU&Exact Match&CodeBLEU\\
				\hline
				Naive copy&18.69&0.0&-&18.54&0.0&-\\
				PBSMT & 40.06& 16.1&43.48&43.53& 12.50&42.71\\
				Transformer & 50.47& 37.90&61.59&55.84& 33.00&63.74\\
				\hline
				RoBERTa (code)& 71.99& 57.90& 80.18 & 77.46& 56.10&83.07\\
				CodeBERT & 72.14 &58.80&79.41 &79.92& 59.00&\textbf{85.10}\\
				GraphCodeBERT & 72.64& 58.80&- & 80.58 &59.40&-\\
				\ourapproach&\textbf{76.52}&\textbf{61.30}&\textbf{82.22}&\textbf{80.75}&\textbf{60.40}&84.85\\
				\hline
		\end{tabular}}
	\end{table*}
	\subsection{Program Translation}
	Table \ref{table:codetrans} shows the experimental results on the program translation task. The baseline results are reported in~\cite{lu2021codexglue} and~\cite{guo2021graphcodebert}. For BERT-like pre-trained models, we leverage them to initialize the encoder, and randomly initialize the parameters of the decoder (6-layer Transformer) and the source-to-target attention weights for the translation task. We can observe that \ourapproach still surpasses all models and achieve the best overall performance. 
	In the C\#$\rightarrow$Java task, \ourapproach outperforms CodeBERT by 4.38 BLEU points, 2.5 exact match points, and 2.81 CodeBLEU points.
	We contend that the reason for the above improvements is the introduction of code syntax knowledge.
	Although \ourapproach has not been pre-trained on the C\#, there is significant syntactic and semantic similarities between C\# and Java. 
	The syntax knowledge integrated into the pre-training phase of \ourapproach can effectively generalize to other programming languages. 
	
	\begin{table}[t!]
		\normalsize
		\centering
		\setlength{\tabcolsep}{1mm}{
			\caption{Ablation study on the natural language code search task evaluating with MRR, using the CodeSearch dataset.}
			\label{table:ablation}
			\begin{tabular}{l|c c c c c c c}
				\hline
				Models & Ruby &JS &Go &PY &Java &PHP &Avg. \\
				\hline
				\ourapproach &\textbf{72.2}&\textbf{67.7}&\textbf{91.3}&\textbf{72.4}&\textbf{72.3}&\textbf{67.8}&\textbf{74.0}\\
				\ \ w/o TEP &72.0&67.5&91.1&72.2&71.9&67.6&73.7\\
				\ \ \ \ w/o IP &71.4&66.7&90.5&71.6&71.2&66.9&73.1\\
				\ \ \ \ \ \ w/o MCL&70.6&64.2&89.3&68.6&68.7&64.6&71.0\\
				\hline
		\end{tabular}}
	\end{table}
	\subsection{Ablation Study}
	In this experiment, we compare several simplified versions of \ourapproach to understand the effect of different components, including AST Edge Prediction (TEP) objective, Identifier Prediction (IP) objective and Multi-modal Contrastive Learning (MCL). 
	As a case study, we take the natural language code search task using the CodeSearch dataset, and present the results in Table~\ref{table:ablation}. The setting of w/o (TEP, IP, MCL) indicates that \ourapproach progressively removes these components.
	We can observe that: (1) the two pre-training objectives (TEP and IP) that exploit the symbolic and syntactic information are effective. Dropping these two components slightly hurts the performance of \ourapproach. (2) MCL plays a more important role for \ourapproach as further cutting this component degrades the performance a lot. MCL may allow the model to maximize the mutual information between different modalities to learn more comprehensive sentence-level representations. This ablation result shows that it contributes a lot to the natural language code search task.
	

	\section{Related Work}
	\paragraph{Pre-Trained Models on Programming Language}
	With the success of pre-trained models in NLP, some recent works attempt to extend the pre-training technologies to codes. The pre-trained models on codes promote the development of code intelligence. \citet{kanade2020learning} developed CuBERT, which is pre-trained on the Python language. They adopted the masked language modeling objective of BERT to obtain general code representations. \citet{feng2020codebert} proposed CodeBERT, which is pre-trained on NL-PL pairs in six programming languages, adding a replaced token detection objective \cite{clark2020electra}. \citet{guo2021graphcodebert} developed GraphCodeBERT based on CodeBERT considering the data flow of codes. Besides the BERT-like models, \citet{svyatkovskiy2020intellicode} and \citet{liu2020multi} respectively proposed CodeGPT and CugLM for code completion task based on the transformer \cite{vaswani2017attention} decoder. 
	\citet{ahmad2021unified} proposed PLBART based on the BART \cite{lewis2020bart} architecture, which is pre-trained on large-scale Java and Python functions paired with natural language comments via denoising autoencoding. \citet{phan2021cotext} proposed CoTexT follows the architecture of T5 \cite{raffel2019exploring}, which employs denoising sequence-to-sequence pre-training on multiple programming languages. 
	\citet{jiang2021treebert} proposed TreeBERT, an encoder-decoder architecture, incorporating the AST into the model. It is pre-trained by tree masked language modeling (TMLM) and node order prediction (NOP) objectives. TreeBERT does not consider the edges in AST, which is not sufficient to exploit rich syntactic structure information within edges.
	Neither of them takes into account the symbolic property of programming languages. Further exploration of the multi-modal potential of programming languages is still insufficient.
	
	\paragraph{Contrastive Learning}
	Contrastive learning has become an emerging field due to its great success in computer vision~\cite{chen2020a, misra2020self, tschannen2020on}. Some works use different data augmentations (spatial/geometric and appearance transformations) to make an image agree with each other, improving the quality of visual representations. Inspired by these, several works try to use contrastive learning on NL and PL.
	\citet{fang2020cert} proposed CERT model, treating back-translated sentence and original sentence as a positive pair. \citet{giorgi2020declutr} presented DeCLUTR model, considering that different spans in the same document are similar to each other. 
	\citet{bui2021self} and  \citet{jain2021contrastive} exploited the contrastive learning on codes. they trained the neural network over a contrastive learning objective to compare similar and dissimilar code snippets. However, they only handle the single modality of code, and ignore the multi-modal characteristic of programming languages.
	
	Multiple modalities contain complementary information that offers the potential for drawing useful connections across different modalities. Some recent works attempt to adopt multi-modal contrastive learning to mine more comprehensive representations \cite{yuan2021multimodal, xu2021contrastive, Hassani2020ContrastiveMR}. \citet{yuan2021multimodal} proposed a method to improve visual representations embracing multi-modal data. They exploited intrinsic data properties within each modality and semantic information from cross-modal correlation simultaneously. \citet{xu2021contrastive} proposed a contrastive multi-modal clustering framework to mine high-level semantic information, considering both multi-modal consistency and diversity. \citet{Hassani2020ContrastiveMR} proposed a multi-view method for learning node and graph level representations by contrasting structural views of graphs.

	\section{Conclusion} 
	In this paper, we have proposed \ourapproach, a syntax-guided multi-modal contrastive pre-training framework for code representation. Considering the symbolic and syntactic property of source code, we design two new pre-training objectives to predict identifiers, and edges between two nodes of AST, respectively.  Meanwhile, to exploit the complementary information in semantically equivalent modalities (i.e., code, comment, AST) of code, We propose a multi-modal contrastive learning strategy to maximize the mutual information among different modalities.
	Comprehensive experiments conducted on four code intelligence tasks demonstrate that \ourapproach achieves state-of-the-art with the same pre-training corpus and model size.
	
	\bibliography{aaai22}

\end{document}